\documentclass[letterpaper]{article}

\usepackage{natbib,alifeconf}  

\usepackage[utf8]{inputenc} 
\usepackage[T1]{fontenc}    
\usepackage{hyperref}       
\usepackage{url}            
\usepackage{booktabs}       
\usepackage{amsfonts}       
\usepackage{nicefrac}       
\usepackage{microtype}      
\usepackage{lipsum}		
\usepackage{graphicx}
\usepackage{subcaption}
\usepackage{graphicx}
\usepackage{xcolor} 
\usepackage{tikz}
\usetikzlibrary{shapes.geometric, arrows, positioning, automata,positioning,fit,backgrounds}
\tikzstyle{process} = [rectangle, minimum width=2cm, minimum height=1cm, text centered, draw=black, fill=white!30]
\tikzstyle{sum} = \tikzstyle{sum} = [draw, circle, minimum size=.5cm]
\tikzstyle{arrow} = [thick,->,>=stealth]

\usepackage{amsmath}

\usepackage{bm}
\usepackage{units}
\usepackage{float}
\usepackage{dblfloatfix}
\usepackage{algorithm}
\usepackage{algpseudocode}
\makeatother
\usepackage[T1]{fontenc}

\usetikzlibrary{fit,calc}
\newcommand*{\tikzmk}[1]{\tikz[remember picture,overlay,] \node (#1) {};\ignorespaces}
\newcommand{\boxit}[1]{\tikz[remember picture,overlay]{\node[yshift=0pt,fill=#1,opacity=.25,fit={(A)($(B)+(.9\linewidth,.5\baselineskip)$)}] {};}\ignorespaces}
\colorlet{yellow}{yellow!100}
\colorlet{blue}{cyan!60}

\title{A Comparative Study of Brain Reproduction Methods for Morphologically Evolving Robots}

\author{Jie Luo$^{1}$, Carlo Longhi$^{2}$ \and Agoston E.
Eiben$^{1}$ \\
\mbox{}\\
$^1$Vrije Universiteit Amsterdam, Amsterdam, The Netherlands \\
$^2$University of Bologna, Bologna, Italy \\
j2.luo@vu.nl} 

\begin{document}
\maketitle

\begin{abstract}
In the most advanced robot evolution systems, both the bodies and the brains of the robots undergo evolution and the brains of `infant' robots are also optimized by a learning process immediately after `birth'. This paper is concerned with the brain evolution mechanism in such a system. In particular, we compare four options obtained by combining asexual or sexual brain reproduction with Darwinian or Lamarckian evolution mechanisms. We conduct experiments in simulation with a system of evolvable modular robots on two different tasks. The results show that sexual reproduction of the robots' brains is preferable in the Darwinian framework, but the effect is the opposite in the Lamarckian system (both using the same infant learning method). Our experiments suggest that the overall best option is asexual reproduction combined with the Lamarckian framework, as it obtains better robots in terms of fitness than the other three. Considering the evolved morphologies, the different brain reproduction methods do not lead to differences. This result indicates that the morphology of the robot is mainly determined by the task and the environment, not by the brain reproduction methods. 

\end{abstract}

\section{Introduction}
In Evolutionary Robotics (ER), the simultaneous development of robot morphologies and control systems  is a difficult task. It was introduced by Karl Sims in his simulated virtual creatures (\cite{sims1994evolving}) and we have only seen relatively simple results so far, as noted by \cite{Cheney2016}. Some of the difficulty is due to the increased dimensionality of the search, but a more pernicious aspect may be the increased ruggedness of the search space: a small variation in the morphology can easily offset the performance of the controller-body combination found earlier. The design choice for reproduction varies per study. \cite{Cheney2016} and \cite{Nygaard2018} employ asexual reproduction for the generation of both morphologies and controllers. \cite{Medvet2021} generate each offspring by either mutation or geometric crossover according to a probability. \cite{Lehman2011}, \cite{miras2020environmental}, \cite{Stensby2021}, \cite{auerbach2012} and \cite{DeCarlo2020} utilize sexual reproduction.

Several researchers have combined evolutionary methods with learning techniques to drive the joint evolution of controllers and morphologies deeper. Different reproduction methods for their robots have been used too. \cite{Cheney2018} produce new offspring by mutation of the parent's morphology or controller but not both. \cite{Wang2019}, \cite{Nygaard2017}, \cite{Kriegman2018} and \cite{Goff2021} only use mutation to generate the robotic offspring and \cite{Gupta2021} generate new morphologies by asexual reproduction but use randomly initialized controllers. Sexual reproduction is used, instead, to produce the morphology and controller of the offspring by \cite{Luo2022}, \cite{Jelisavcic2019}, and \cite{miras2018effects}.

To the best of our knowledge, no studies have addressed the arity of brain reproduction (a.k.a. crossover or recombination), comparing unary/asexual with binary/sexual reproduction in a morphologically evolving robotic system, let alone in combination with a Darwinian vs. Lamarckian framework for combining evolution with learning. Our study aims to fill this gap by answering the following research questions:

\textbf{Research Question 1:} How do asexual and sexual brain reproduction compare within Lamarckian and Darwinian evolution frameworks in terms of task performance?


\textbf{Research Question 2:} Will the different brain reproduction methods lead to different robot morphologies? 


\textbf{Research Question 3:} Will the different brain reproduction methods lead to different robot behaviours?

To answer these questions, we design an evolutionary robot system in which robot morphologies (bodies) and controllers (brain) are jointly evolved and, before reproduction, the controllers go through a phase of learning. In such systems, the best robots are chosen to produce new robotic offspring and the new genes are formed by sexual or asexual reproduction of their parent's genes. The body of every offspring is produced by recombination and mutation of the genotypes of its parents' bodies. For the inheritance of the brain, we design two different mechanisms. The first one, dubbed sexual reproduction, generates the new brain by recombination and mutation starting from the parent's genotype. The second mechanism, asexual reproduction, produces the new brain by the sole mutation of the genotype of its best parent. The new robots are then tested on two separate tasks, panoramic rotation and point navigation, and a fitness value is assigned to them based on their performance. The best individuals are selected to form the new population.

\section{Methods}
\subsection{Robot Morphology(Body)}
\subsubsection{Body Phenotype}
We choose RoboGen's components as the phenotype of the robot body. RoboGen \cite{Auerbach2014} is a widely used, open-source platform for the evolution of robots which provides a set of modular components: a morphology consists of one core component, one or more brick components, and one or more active hinges. The phenotype follows a tree structure, with the core module being the root node from which further components branch out. Child modules can be rotated 90 degrees when connected to their parent, making 3D morphologies possible. 

\subsubsection{Body Genotype}
The phenotype of bodies is encoded in a Compositional Pattern Producing Network (CPPN) which was introduced by Stanley \cite{Stanley2007} and has been successfully applied to the evolution of both 2D and 3D robot morphologies in prior studies. The structure of the CPPN has four inputs and five outputs. The first three inputs are the x, y, and z coordinates of a component, and the fourth input is the distance from that component to the core component in the tree structure. The first three outputs are the probabilities of the modules being a brick, a joint, or empty space, and the last two outputs are the probabilities of the module being rotated 0 or 90 degrees. For both module type and rotation the output with the highest probability is always chosen; randomness is not involved.

The body's genotype to phenotype decoder operates as follows:\\
The core component is generated at the origin. We move outwards from the core component until there are no open sockets(breadth-first exploration), querying the CPPN network to determine the type and rotation of each module. Additionally, we stop when ten modules have been created. The coordinates of each module are integers; a module attached to the front of the core module will have coordinates (0,1,0). If a module would be placed on a location already occupied by a previous module, the module is simply not placed and the branch ends there. In the evolutionary loop for generating the body of offspring, we use the same mutation and crossover operators as in MultiNEAT (\url{https://github.com/MultiNEAT/}).

\subsection{Robot Controller(Brain)}
\subsubsection{Brain Phenotype}
\begin{figure}
    \centering
    \includegraphics[width=0.9\linewidth]{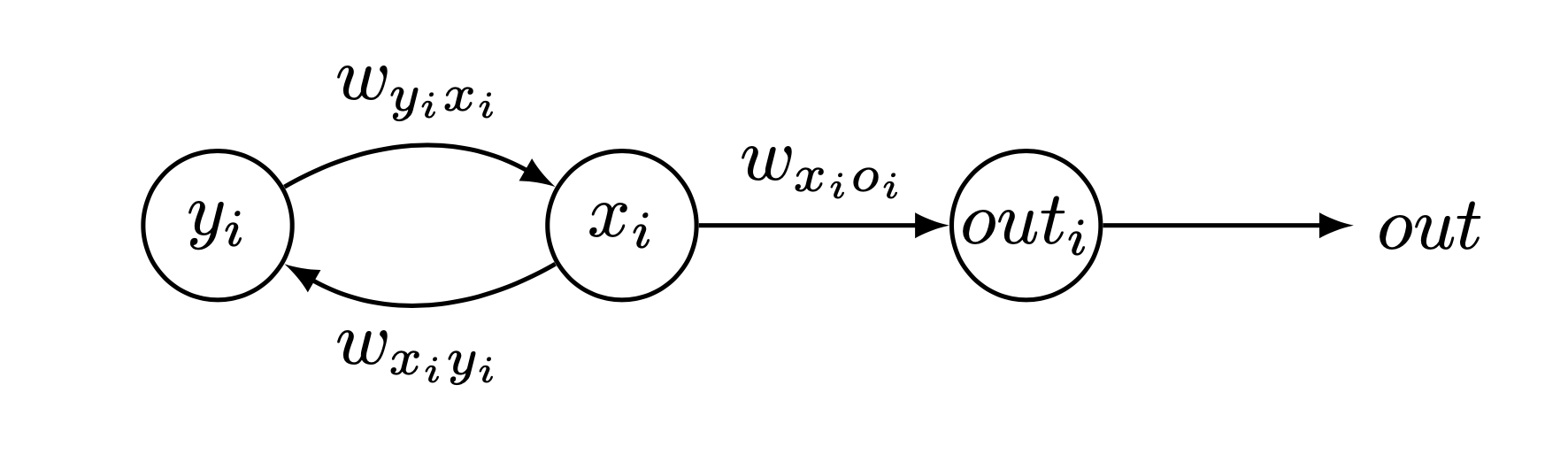}
    \caption{The structure of the CPG associated to the $i_{th}$ joint. $w_{x_iy_i}$, $w_{y_ix_i}$ and $w_{x_io_i}$ are the weights of the connections between the neurons and out is the activation value of $out_i$ neuron that controls the servo in a joint}
    \label{fig:cpg}
\end{figure}

We use Central Pattern Generators (CPGs)-based controllers to drive the modular robots which has demonstrated their success in controlling various types of robots, from legged to wheeled ones in previous research. Each joint of the robot has an associated CPG that is defined by three neurons: an $x_i$-neuron, a $y_i$-neuron and an $out_i$-neuron. The recursive connection of the tree neurons is shown in Figure \ref{fig:cpg}. 
The change of the $x_i$ and $y_i$ neurons' states with respect to time is obtained by multiplying the activation value of the opposite neuron with the corresponding weight  $\dot{x}_i = w_i y_i$, $\dot{y}_i = -w_i x_i$. To reduce the search space we set $w_{x_iy_i}$ to be equal to $-w_{y_ix_i}$ and call their absolute value $w_i$. The resulting activations of neurons $x_i$ and $y_i$ are periodic and bounded. The initial states of all $x$ and $y$ neurons are set to $\frac{\sqrt{2}}{2}$ because this leads to a sine wave with amplitude 1, which matches the limited rotating angle of the joints.

\begin{figure}
   \begin{minipage}{0.49\textwidth}
     \centering
     \includegraphics[width=.9\linewidth]{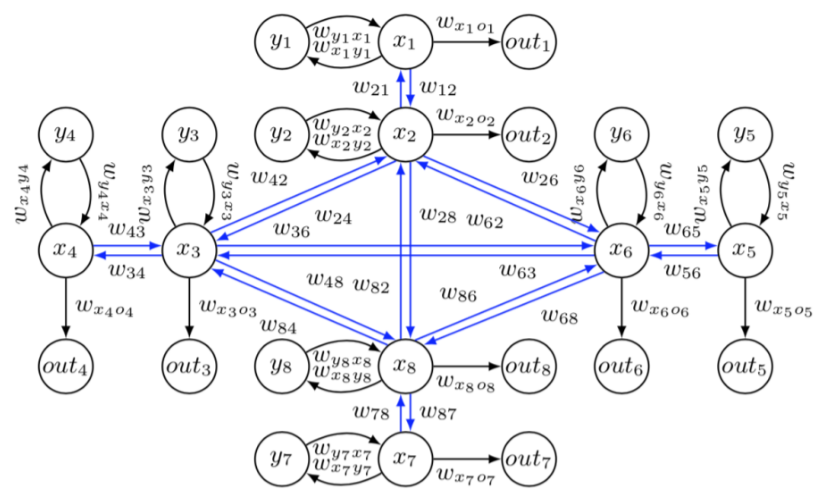}
    \end{minipage}
   \centering
    \caption{\label{fig:cpg_network}Brain phenotype (CPG network) of a "+" shape robot. In our design, the topology of the brain is determined by the topology of the body.}
\end{figure}

To enable more complex output patterns, connections between CPGs of neighbouring joints are implemented. An example of the CPG network of a "+" shape robot is shown in Figure \ref{fig:cpg_network}. Two joints are said to be neighbours if their distance in the morphology tree is less than or equal to two. 
Consider the $i_{th}$ joint, and $\mathcal{N}_i$ the set of indices of the joints neighbouring it, $w_{ij}$ the weight of the connection between $x_i$ and $x_j$. Again, $w_{ij}$ is set to be $-w_{ji}$. The extended system of differential equations becomes:

\begin{equation}
    \begin{split}
        \dot{x}_i &= w_i y_i + \sum_{j \in \mathcal{N}_i} w_{ji} x_j \\
        \dot{y}_i &= -w_i x_i
    \end{split}
\end{equation}

Because of this addition, $x$ neurons are no longer bounded between $[-1,1]$. For this reason, we use the hyperbolic tangent function (\emph{tanh}) as the activation function of $out_i$-neurons.

\begin{equation}
    out_{(i,t)}(x_{(i,t)}) = \frac{2}{1+e^{-2x_{(i,t)}}} - 1
\end{equation}

\subsubsection{Brain Genotype}
We utilize an array-based structure for the brain's genotypic representation to map the CPG weights. This is achieved via direct encoding, a method chosen specifically for its potential to enable reversible encoding in future stages.
We have seen how every modular robot can be represented as a 3D grid in which the core module occupies the central position and each module's position is given by a triple of coordinates. When building the controller from our genotype, we use the coordinates of the joints in the grid to locate the corresponding CPG weight. To reduce the size of our genotype, instead of the 3D grid, we use a simplified 3D in which the third dimension is removed. For this reason, some joints might end up with the same coordinates and will be dealt with accordingly. 

Since our robots have a maximum of 10 modules, every robot configuration can be represented in a grid of $21 \times 21$. Each joint in a robot can occupy any position of the grid except the center. For this reason, the possible positions of a joint in our morphologies are exactly $(21 \cdot 21) - 1=440$. We can represent all the internal weights of every possible CPG in our morphologies as a $440$-long array. When building the phenotype from this array, we can simply retrieve the corresponding weight starting from a joint's coordinates in the body grid.

To represent the external connections between CPGs, we need to consider all the possible neighbours a joint can have. In the 2-dimensional grid, the number of cells in a distance-2 neighbourhood for each position is represented by the Delannoy number $D(2,2) = 13$, including the central element. Each one of the neighbours can be identified using the relative position from the joint taken into consideration. Since our robots can assume a 3D position, we need to consider an additional connection for modules with the same 2D coordinates.

\begin{figure}
    \centering
    \includegraphics[width=0.47\textwidth]{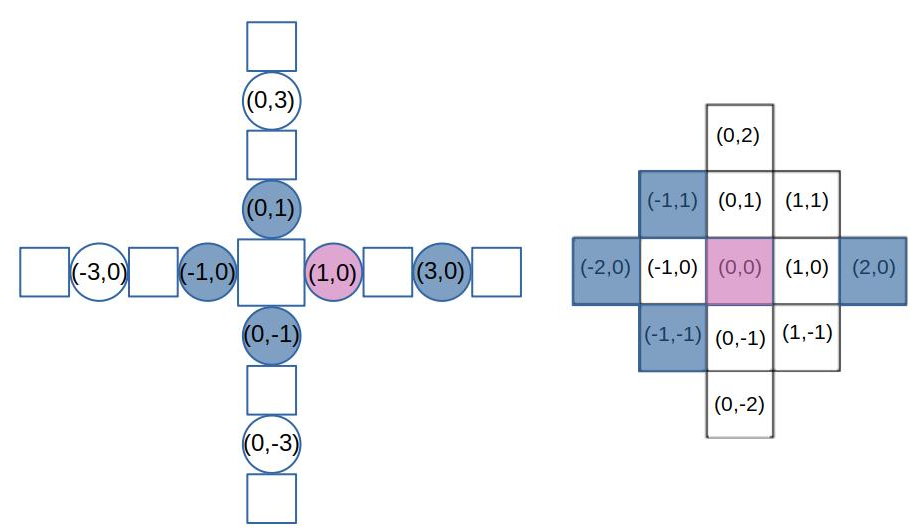}
    \caption{Brain genotype to phenotype mapping of a "+" shape robot. The left image (brain phenotype)  shows the schema of the "+" shape robot with the coordinates of its joints in the 2D body grid. The right image (brain genotype) is the distance 2 neighbour of the joint at (1,0). The coordinates reported in the neighbourhood are relative to this joint. The CPG weight of the joint is highlighted in purple and its 2-distance neighbours are in blue.}
    \label{fig:brain_geno}
\end{figure}

To conclude, for each of the $440$ possible joints in the body grid, we need to store 1 internal weight for its CPG, 12 weights for external connections, and 1 weight for connections with CPGs at the same coordinate for a total of 14 weights. The genotype used to represent the robots' brains is an array of size $440 \times 14$. An example of the brain genotype of a "+" shape robot is shown in Figure \ref{fig:brain_geno}.

It is important to notice that not all the elements of the genotype matrix are going to be used by each robot. This means that their brain's genotype can carry additional information that could be exploited by their children with different morphologies.

The recombination operator for the brain genotype is implemented as a uniform crossover where each gene is chosen from either parent with equal probability. The new genotype is generated by essentially flipping a coin for each element of the parents' genotype to decide whether or not it will be included in the offspring's genotype. In the uniform crossover operator, each gene is treated separately.
The mutation operator applies a Gaussian mutation to each element of the genotype by adding a value, with a probability of 0.8, sampled from a Gaussian distribution with 0 mean and 0.5 standard deviation.

\subsection{Asexual \& Sexual Reproduction}
In this research, the bodies of the robots are evolved only with sexual reproduction while the brains of the robots are evolved with asexual or sexual reproduction.

Body - sexual reproduction: Parents are selected from the current generation using binary tournaments with replacement. We perform two tournaments in which two random potential parents each are selected. In each tournament the potential parents are compared, the one with the highest fitness wins the tournament and becomes a parent. The body of every new offspring is created through recombination and mutation of the genotypes of its parents.

Brain - asexual \& sexual reproduction: For the generation of the brain, we use two different strategies. The first strategy is called asexual because the brain genotype of the offspring is generated from only one parent. The brain genotype of the best-performing parent is mutated before being inherited by its offspring. For sexual reproduction, instead, the child's brain is created through the recombination and mutation of its parents' brain genotypes.

The Algorithm \ref{alg:EL} displays the pseudocode of the complete integrated process of evolution and learning. With the highlighted blue code, it is the sexual reproduction method, without it is the asexual reproduction. With the highlighted yellow code, it is the Lamarckian learning mechanism, without it is the Darwinian learning mechanism. Note that for the sake of generality, we distinguish two types of quality testing depending on the context, evolution or learning. Within the evolutionary cycle (line 2 and line 14) a test is called an evaluation and it delivers a fitness value. Inside the learning cycle (line 11) a test is called an assessment and it delivers a performance value. 


The code for replicating this work and carrying out the experiments is available online: \url{https://rb.gy/7gx13}. 

\begin{algorithm}[h!]
  \caption{Evolution+Learning}
  \label{alg:EL}
  \begin{algorithmic}[1]
    \State INITIALIZE robot population 
    \State EVALUATE each robot 
    \While{not STOP-EVOLUTION}
        \State SELECT parents; 
        \State RECOMBINE+MUTATE parents' bodies; 
        
        \tikzmk{A}
        \State RECOMBINE parents' brains;
        
        \tikzmk{B} \boxit{blue}
        \State MUTATE parents' brains; 
        \State CREATE offspring robot body; 
        \State CREATE offspring robot brain; 
        
        \State INITIALIZE brain(s) for the learning process; 
        \While{not STOP-LEARNING}
            \State ASSESS offspring; 
            \State GENERATE new brain for offspring;
        \EndWhile 
        \State EVALUATE offspring with the learned brain; 
        
        \tikzmk{A}
        \State UPDATE brain genotype 

        \tikzmk{B} \boxit{yellow}
        \State SELECT survivors / UPDATE population
          
    \EndWhile
 \end{algorithmic}
\end{algorithm}

\subsection{Learning algorithm}
We use Reversible Differential Evolution (RevDE) \cite{Tomczak2020} as the learning algorithm because it has proven to be effective in previous research \cite{Luo2022}. This method works as follows:
\begin{enumerate}
    \item Initialize a population with \textit{$\mu$} samples ($n$-dimensional vectors), $\mathcal{P}_{\mu}$. 
    \item Evaluate all \textit{$\mu$} samples.
    \item Apply the reversible differential mutation operator and the uniform crossover operator.\\
    \textit{The reversible differential mutation operator}: Three new candidates are generated by randomly picking a triplet from the population, $(\mathbf{w}_i,\mathbf{w}_j,\mathbf{w}_k)\in \mathcal{P}_{\mu}$, then all three individuals are perturbed by adding a scaled difference in the following manner:
        \begin{equation}\label{eq:de3}
            \begin{split}
            \mathbf{v}_1 &= \mathbf{w}_i + F \cdot (\mathbf{w}_j-\mathbf{w}_k) \\
            \mathbf{v}_2 &= \mathbf{w}_j + F \cdot (\mathbf{w}_k-\mathbf{v}_1) \\
            \mathbf{v}_3 &= \mathbf{w}_k + F\cdot (\mathbf{v}_1-\mathbf{v}_2) 
            \end{split}
        \end{equation}
        where $F\in R_+$ is the scaling factor. New candidates $y_1$ and $y_2$ are used to calculate perturbations using points outside the population. This approach does not follow the typical construction of an EA where only evaluated candidates are mutated.\\
        \textit{The uniform crossover operator}: Following the original DE method \cite{Storn1997}, we first sample a binary mask $\mathbf{m} \in \{0, 1\}^D$ according to the Bernoulli distribution with probability \textit{$CR$} shared across $D$ dimensions, and calculate the final candidate according to the following formula:
        \begin{equation}\label{eq:de2}
              \mathbf{u} = \mathbf{m} \odot \mathbf{w}_n+(1-m) \odot \mathbf{w}_n .
        \end{equation}
        Following general recommendations in literature \cite{Pedersen2010} to obtain stable exploration behaviour, the crossover probability CR is fixed to a value of $0.9$ and according to the analysis provided in \cite{Tomczak2020}, the scaling factor $F$ is fixed to a value of 0.5. 
    \item Perform a selection over the population based on the fitness value and select \textit{$\mu$} samples.
    \item Repeat from step (2) until the maximum number of iterations is reached.
\end{enumerate}

As explained above, we apply RevDE here as a learning method for `newborn' robots. In particular, it will be used to optimize the weights of the CPGs of our modular robots for the tasks during the Infancy stage. 
\subsection{Tasks and Fitness functions}
\paragraph{Point Navigation}
Point navigation is a closed-loop controller task which needs feedback (coordinates)from the environment passing to the controller to steer the robot. The coordinates are used to obtain the angle between the current position and the target. If the target is on the right, the right joints are slowed down and vice versa. 

A robot is spawned at the centre of a flat arena (10 × 10 m2) to reach a sequence of target points $P_1,..., P_N$. In each evaluation, the robot has to reach as many targets in order as possible. Success in this task requires the ability to move fast to reach one target and then quickly change direction to another target in a short duration. A target point is considered to be reached if the robot gets within 0.01 meters from it. Considering the experimental time, we set the simulation time per evaluation to be 40 seconds which allows robots to reach at least 2 targets $P_1(1,-1), P_2(0,-2)$.

The data collected from the simulator is the following:
\begin{itemize}
    \item The coordinates of the core component of the robot at the start of the simulation are approximate to $P_0 (0,0)$;
    \item The coordinates of the robot, sampled during the simulation at 5Hz, allowing us to plot and approximate the length of the followed path;
    \item The coordinates of the robot at the end of the simulation $P_T(x_T,y_T)$;
    \item The coordinates of the target points $P_1(x_1,y_1)$... $P_n(x_n,y_n)$.
    \item The coordinates of the robot, sampled during the simulation at 5Hz, allow us to plot and approximate the length of the path $L$.
\end{itemize}

The fitness function for this task is designed to maximize the number of targets reached and minimize the path followed by the robot to reach the targets.
\begin{multline}
    F=\sum_{i=1}^{k}dist(P_i,P_{i-1}) \\
    +(dist(P_k,P_{k-1}) - dist(P_T,P_k)) \\
    - \omega \cdot L
\end{multline}
where $k$ is the number of target points reached by the robot at the end of the evaluation, and $L$ is the path travelled. The first term of the function is a sum of the distances between the target points the robot has reached. The second term is necessary when the robot has not reached all the targets and it calculates the distance travelled toward the next unreached target. The last term is used to penalize longer paths and $\omega$ is a constant scalar that is set to 0.1 in the experiments. E.g., if a robot just reached 2 targets, the maximum fitness value will be $dist(P_1, P_0)+(dist(P_2, P_1)-dist(P2, P2))-0.1*L=\sqrt{2}+\sqrt{2}-0.2*\sqrt{2} \approx 2.54$ ($L$ is shortest path length to go through $P_1$ and $P_2$ which is equal to $2*\sqrt{2}$).

\paragraph{Panoramic Rotation}
The panoramic Rotation task is an open-loop controller task which does not need any feedback from the environment to feed the controller. Same as the Point navigation, the initial coordinate of the robot is [0,0]. Success in this task requires the ability to rotate 360 degrees around the robot's vertical axis as many times as possible in the evaluation time. 

To solve this task, we collect from the simulator the orientation of the robot which is represented as quaternions sampled at 5 Hz during the evaluation. 

The fitness function is the total rotation (in radians) of the robot computed as the sum of the rotation of the orientation vector at each consecutive timestamp. Simulation time is set to 30 seconds for this task.
\begin{equation}
    F = \sum_{n=1}^{30} \theta_i
\end{equation}
where $\theta_i$ is the angle between two vectors which were converted from quaternions.

We assign a positive sign to the counter-clockwise rotations and a negative one to the clockwise rotations in our tests.
\section{Experimental setup}
The stochastic nature of evolutionary algorithms requires multiple runs under the same conditions and a sound statistical analysis (\cite{bartz2007experimental}). We perform 10 runs for each evaluation task, reproduction mechanism and evolutionary framework, namely Rotation Asexual Darwinian, Rotation Asexual Lamarckian, Rotation Sexual Darwinian, Rotation Sexual Lamarckian, Point Navigation Asexual Darwinian, Point Navigation Asexual Lamarckian, Point Navigation Sexual Darwinian, Point Navigation Sexual Lamarckian. In total, 80 experiments.

Each experiment consists of 30 generations with a population size of 50 individuals and 25 offspring. A total of $50+(25\cdot(30-1))=775$ morphologies and controllers are generated, and then the learning algorithm RevDE is applied to each controller. For RevDE we use a population of 10 controllers for 10 generations, for a total of $(10+30\cdot(10-1))=280$ performance assessments.

The fitness measures used to guide the evolutionary process are the same as the performance measure used in the learning loop. For this reason, we use the same test process for both.
The tests for the task of point navigation use 40 seconds of evaluation time with two target points at the coordinates of $(1, -1)$ and $(0, -2)$. The evaluation time for panoramic rotation is 30 seconds.

All the experiments are run with Mujoco simulator-based wrapper called Revolve2 on a 64-core Linux computer, where they each take approximately 15 hours to finish, totalling 1,200 hours of computing time.

\section{Results}
To compare the effects of asexual and sexual reproduction, we consider two generic performance indicators: efficiency and efficacy, meanwhile we also look into robots' behaviour and morphologies.
\subsubsection{Efficacy}
We measure efficacy by the mean and maximum fitness value within the simulation time at the end of the evolutionary process (30 generations), and then we take the average over 10 independent repetitions. 

\begin{figure*}[ht!] 
  \centering
     \begin{subfigure}[b]{0.49\textwidth}
         \centering
         \includegraphics[width=0.75\textwidth]{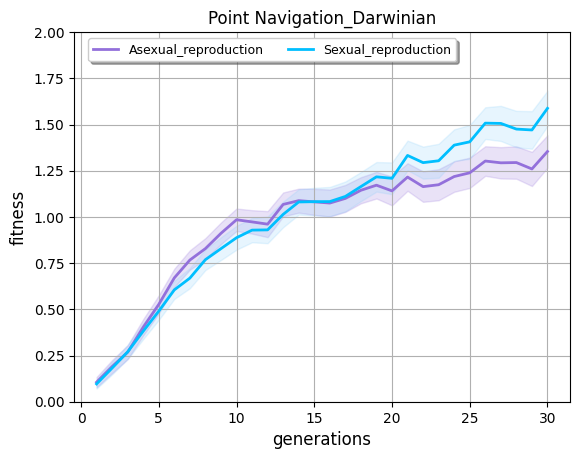}
         \caption{}
     \end{subfigure}
     \hfill
     \begin{subfigure}[b]{0.49\textwidth}
         \centering
         \includegraphics[width=0.75\textwidth]{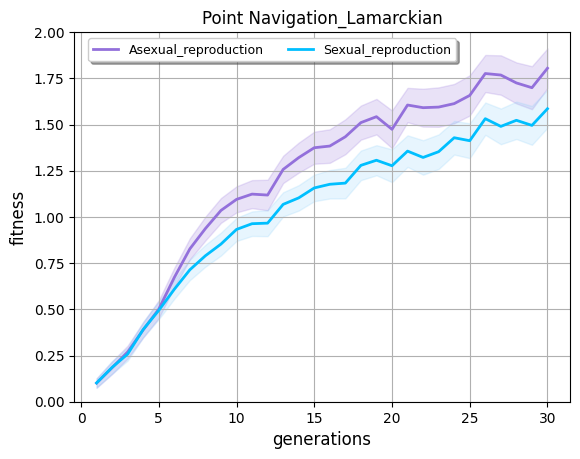}
         \caption{}
     \end{subfigure}
     \hfill
     \begin{subfigure}[b]{0.49\textwidth}
         \centering
         \includegraphics[width=0.75\textwidth]{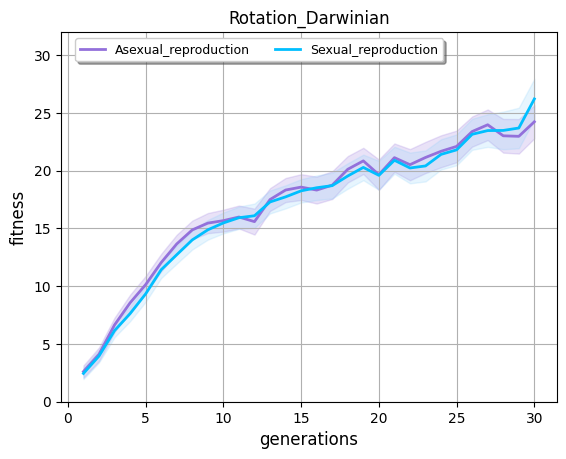}
         \caption{}
     \end{subfigure}
     \hfill
     \begin{subfigure}[b]{0.49\textwidth}
         \centering
         \includegraphics[width=0.75\textwidth]{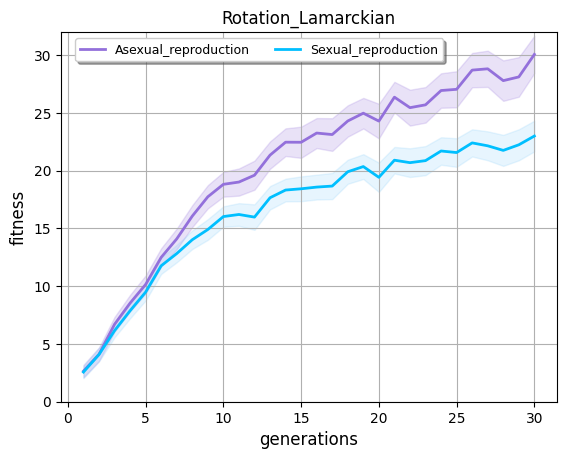}
         \caption{}
     \end{subfigure}
     \hfill
  \caption{Mean fitness over 30 generations (averaged over 10 runs) for asexual reproduction in purple and sexual reproduction in blue. Subfigures (a)(b) exhibit mean average fitness for the point navigation task, and Subfigures (c)(d) are for the rotation task. The bands indicate the 95\% confidence intervals ($\pm1.96\times SE$, Standard Error).}
  \label{fig:fitness_mean_avg} 
\end{figure*}

Figure \ref{fig:fitness_mean_avg} shows both reproduction methods can generate robots that can solve the two tasks successfully. It also shows that asexual and sexual reproduction has the opposite effect for different evolution frameworks on two tasks. With the point navigation task, sexual reproduction achieved a higher fitness value in the Darwinian framework, while in the Lamarckian framework, asexual reproduction has a higher mean fitness value across generations. 

With the panoramic rotation task, there is no significant difference between the two reproduction methods in the Darwinian framework, however, in the Lamarckian framework, the asexual method has a significantly higher mean fitness value across generations than the sexual method.

We can indicate from Figure \ref{fig:fitness_mean_avg} that Lamarckian benefits more without crossover of the learned brain genomes while Darwinian which does not inherit the learned traits is not affected so much with or without crossover for the brain genomes.

\begin{figure*}[ht!] 
  \centering
     \begin{subfigure}[b]{0.49\textwidth}
         \centering
         \includegraphics[width=0.94\textwidth]{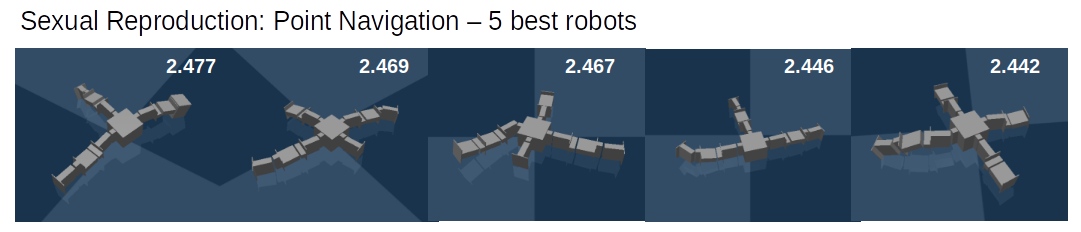}
         \caption{}
     \end{subfigure}
     \hfill
     \begin{subfigure}[b]{0.49\textwidth}
         \centering
         \includegraphics[width=0.94\textwidth]{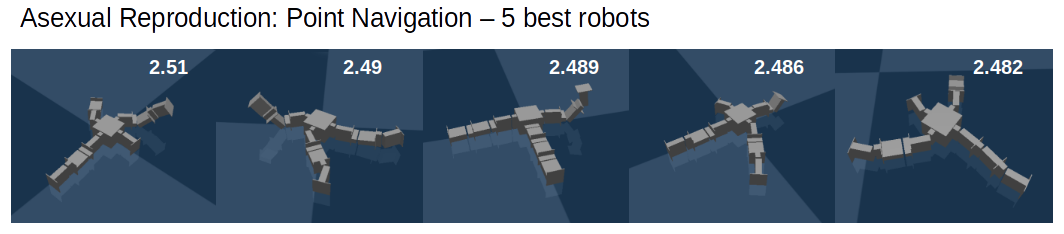}
         \caption{}
     \end{subfigure}
     \hfill
     \begin{subfigure}[b]{0.49\textwidth}
         \centering
         \includegraphics[width=0.94\textwidth]{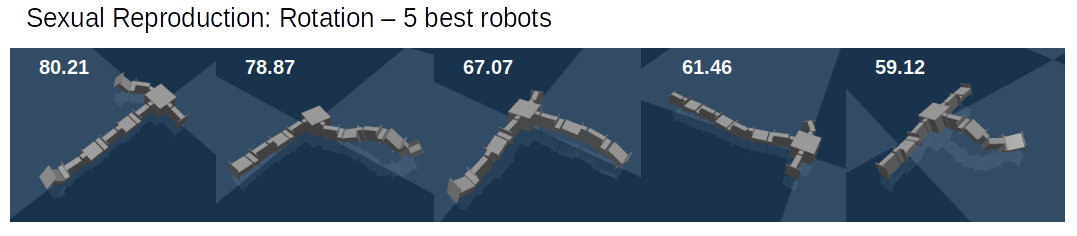}
         \caption{}
     \end{subfigure}
     \hfill
     \begin{subfigure}[b]{0.49\textwidth}
         \centering
         \includegraphics[width=0.94\textwidth]{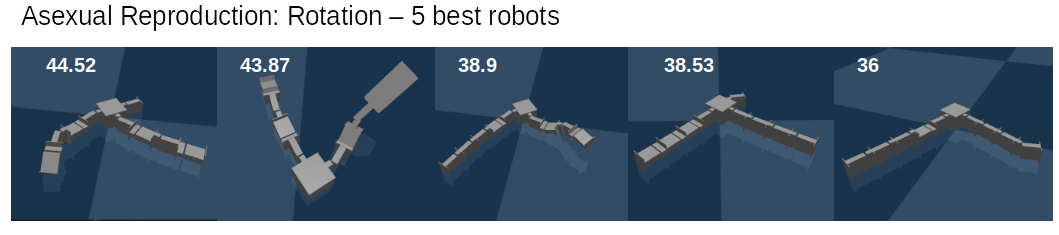}
         \caption{}
     \end{subfigure}
     \hfill
  \caption{Subfigures (a)(b): The 5 best robots produced for the point navigation task by sexual and asexual reproduction and their fitnesses. Sexually reproduced robots have higher fitness values. Subfigures (c)(d): The 5 best robots produced for the panoramic rotation task by sexual and asexual reproduction and their fitnesses. Sexually reproduced robots have higher fitness values and are more diverse.}
  \label{fig:best5} \vspace{-1em}
\end{figure*}
\begin{figure*}
   \begin{minipage}{0.48\textwidth}
     \centering
     \includegraphics[width=0.8\linewidth]{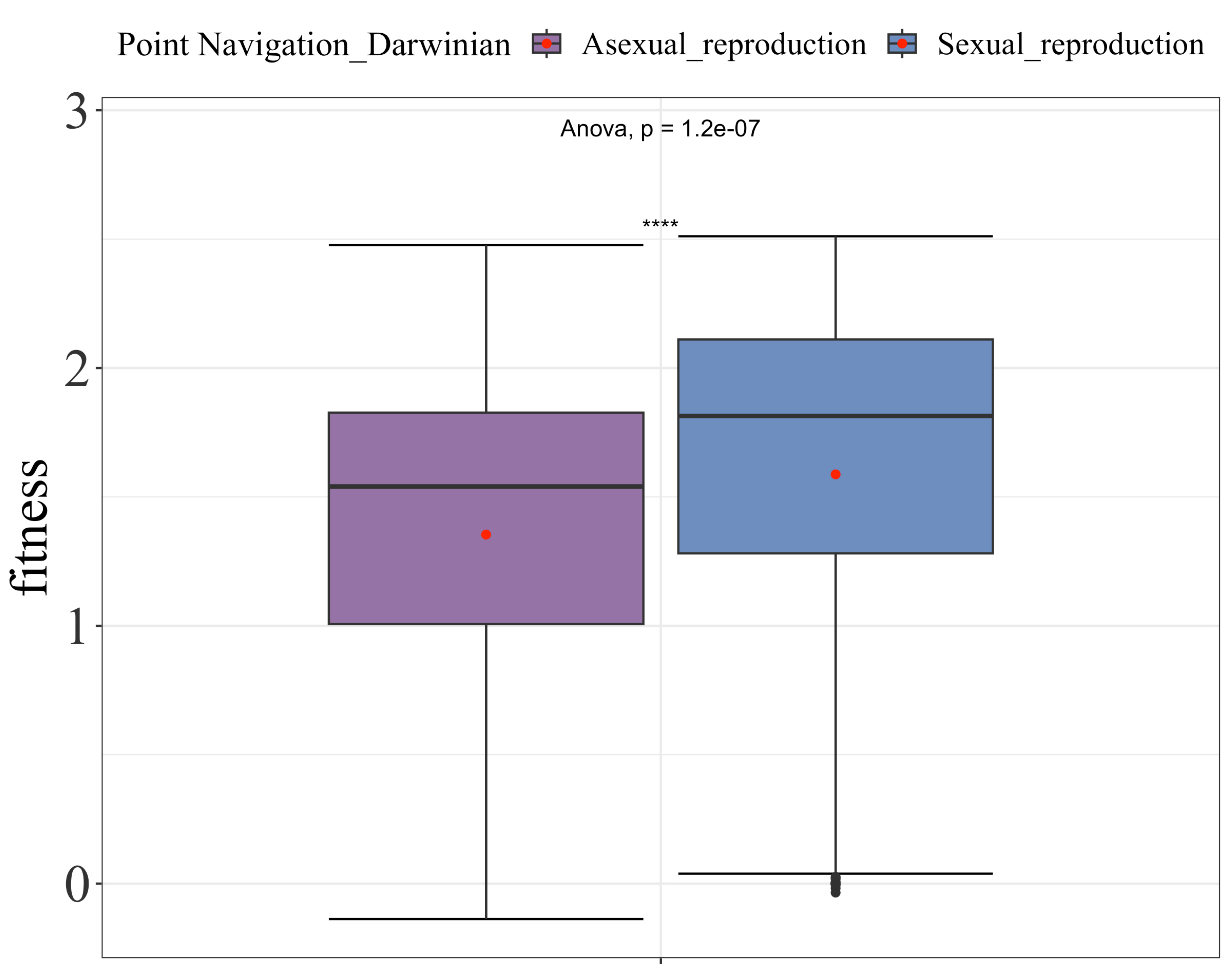}
   \end{minipage}\hfill
   \begin{minipage}{0.48\textwidth}
     \centering
     \includegraphics[width=.8\linewidth]{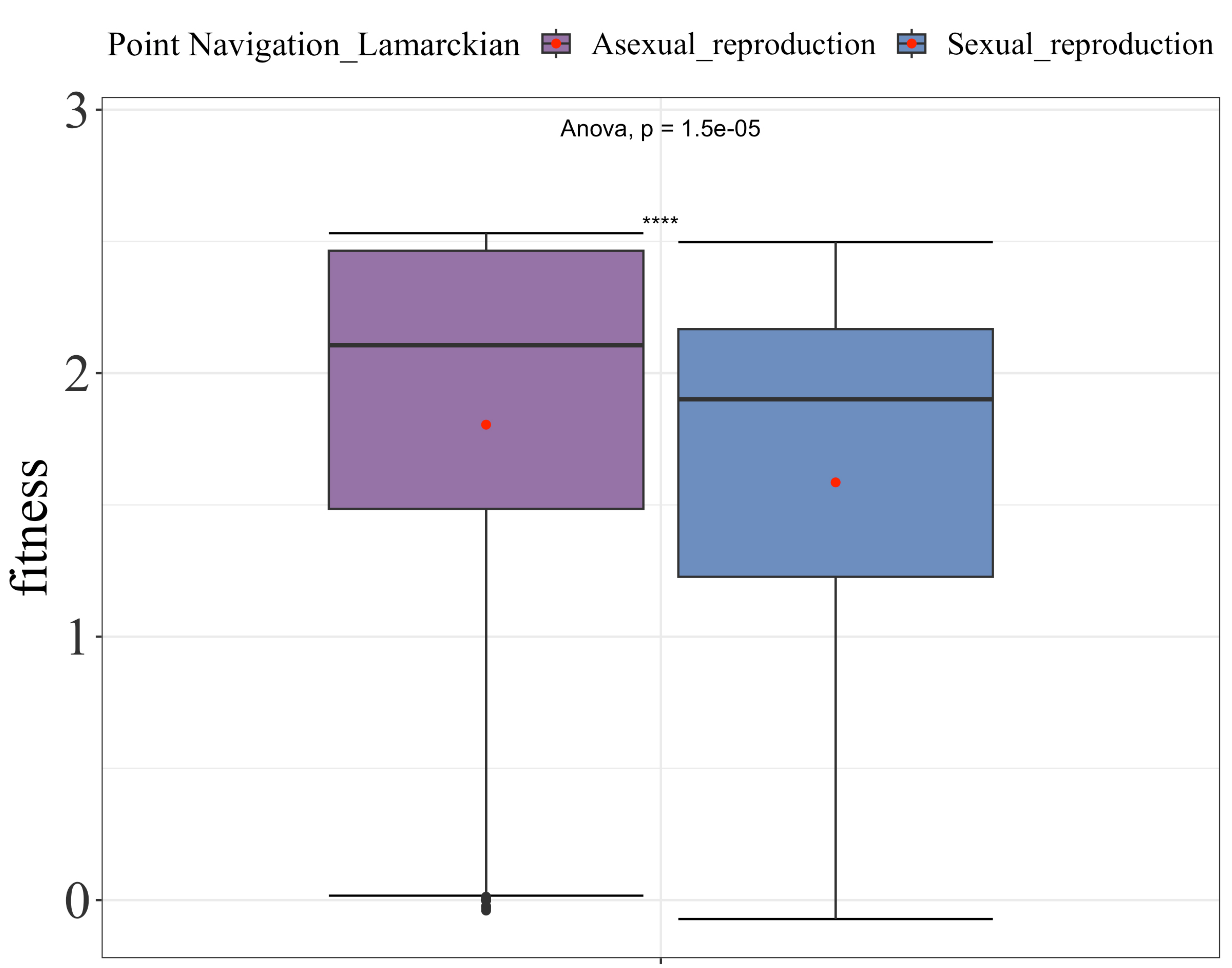}
    \end{minipage}
       \begin{minipage}{0.48\textwidth}
     \centering
     \includegraphics[width=0.82\linewidth]{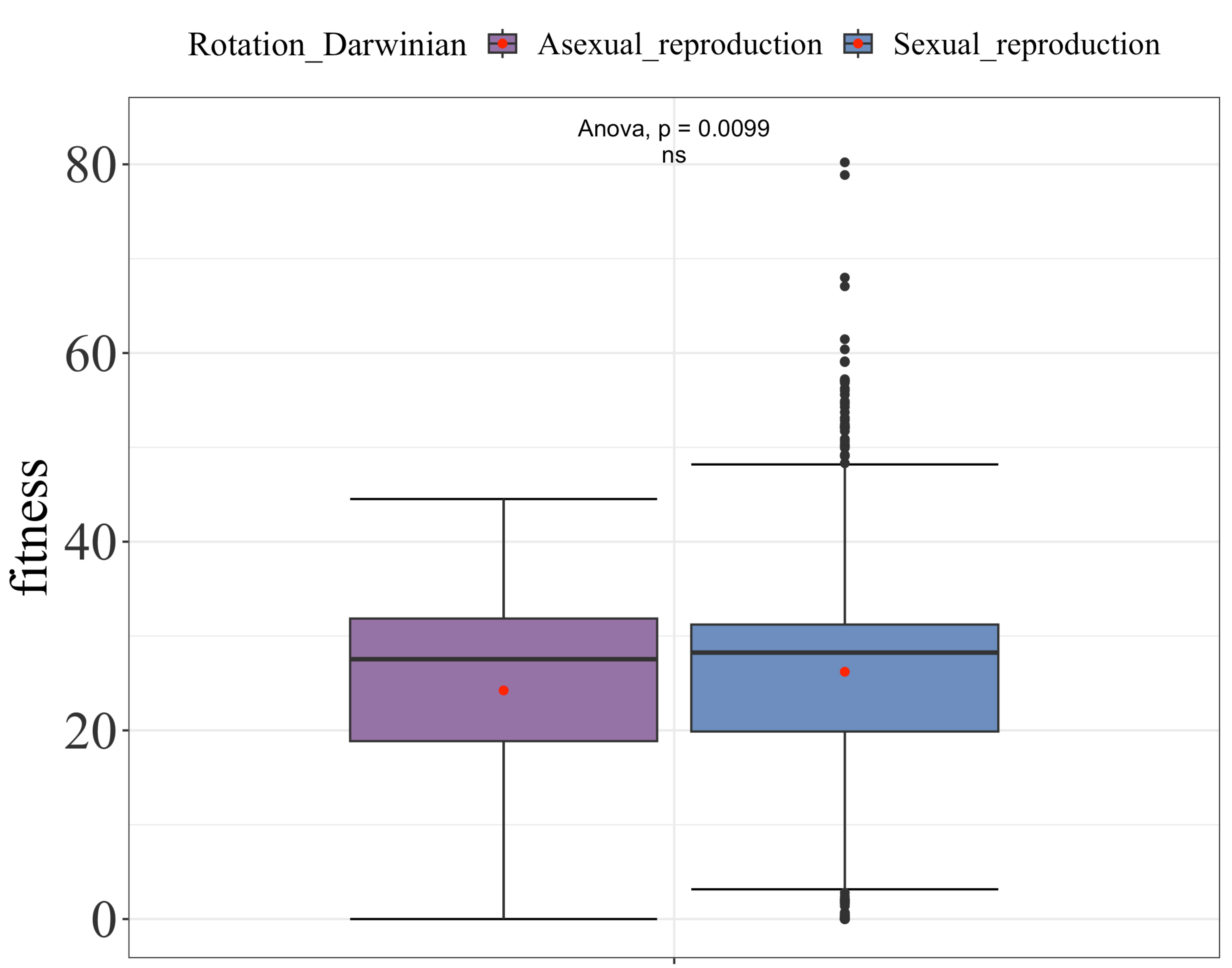}
   \end{minipage}\hfill
   \begin{minipage}{0.48\textwidth}
     \centering
     \includegraphics[width=.82\linewidth]{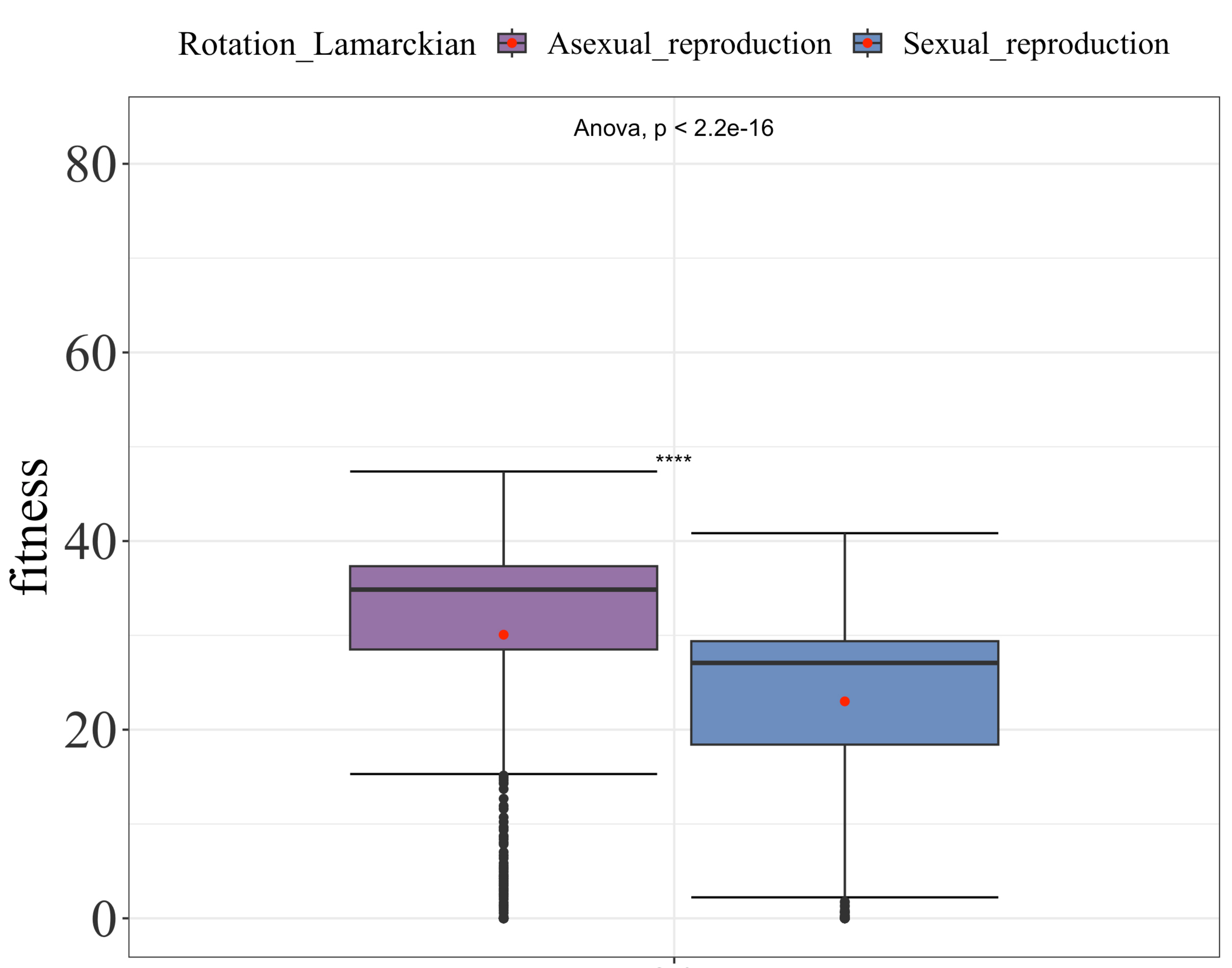}
    \end{minipage}
   \centering
    \caption{\label{fig:efficacy_boxplot} Efficacy boxplots at the generation 30. The fitness values of two reproduction methods at generation 30 for two tasks and two evolution frameworks. Red dots show mean values.}
\end{figure*}

Second, another way to measure the efficacy of the solution is by giving the same computational budget (number of generation) and measuring which method finds the best solution (maximum fitness) faster. In Figure \ref{fig:efficacy_boxplot}, for the Darwinian framework, sexual reproduction has a better mean fitness on point navigation task. Although the difference in mean fitness is not significant on the rotation task, the best sexually reproducing robot for the rotation task with the highest fitness value, 80.21, is almost twice as better as the best robot by asexual reproduction, whose fitness is only 44.52. For the Lamarckian framework, mean and max fitness values of asexual reproduction are significantly better than sexual reproduction's on both tasks.

\begin{figure*}
    \centering
    \begin{subfigure}[b]{0.485\textwidth}
        \centering
        \includegraphics[width=0.9\textwidth]{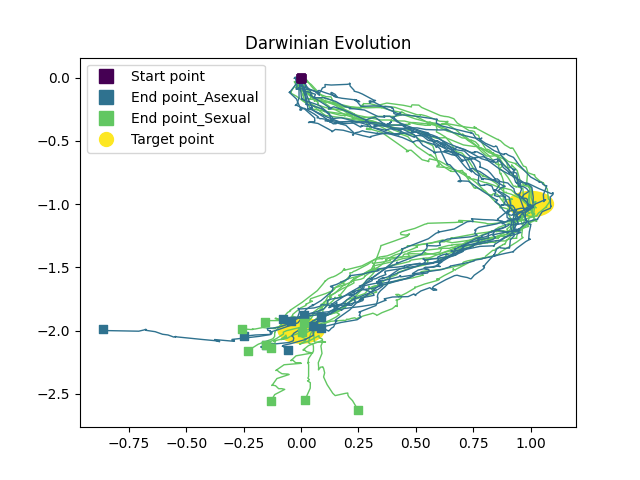}
    \end{subfigure}
    \hfill
    \begin{subfigure}[b]{0.485\textwidth}  
        \centering 
        \includegraphics[width=0.9\textwidth]{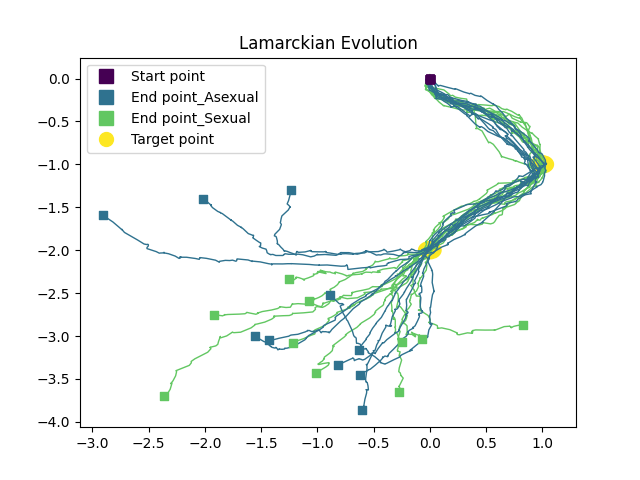}
    \end{subfigure}
    \caption{Trajectories of the best 10 robots from both reproduction methods in the point navigation task with Darwinian and Lamarckian evolution frameworks. The purple square is the starting point. Two yellow circles are the target points which robots aim to go through. The blue lines are the trajectory paths of robots produced by the asexual reproduction method ending at the blue squares. The green lines are from the sexual reproduction method ending at the green squares.}
    \label{fig:trajectories}
\end{figure*}

\subsubsection{Efficiency}
Efficiency indicates how much effort is needed to reach a given quality threshold (the fitness level). In this paper, we use the average number of evaluations to solution to measure it. In figure \ref{fig:fitness_mean_avg}, the most efficient method for the point navigation task is asexual reproduction with the Lamarckian framework. It already surpassed in its 15, 21, and 21 generations the fitness levels that asexual reproduction with Darwinian, sexual reproduction with Darwinian and sexual reproduction with Lamarckian methods achieved at the end of the evolutionary period respectively. So does the rotation task, asexual reproduction with Lamarckian framework surpassed in its 16, 17, and 25,  generations the fitness levels that sexual with Lamarckian method, asexual with Darwinian, and sexual with Darwinian achieved at the end of the evolutionary period respectively.

\subsubsection{Robot Behavior}
To get a better understanding of the robots' behaviour, we visualize the trajectories of the 10 best-performing robots from both reproduction methods for the point navigation task in the last generation across all runs. Figure\ref{fig:trajectories} shows that with the Lamarckian framework, all the robots reached the two target points much earlier than the ones with the Darwinian framework. In the Darwinian framework, both reproduction methods reached the first target point successfully. However, the majority of the robots that use sexual reproduction with the Darwinian framework reach the second target point while only some of the robots produced using asexual reproduction can do the same. In the Lamarckian framework, both reproduction methods reached two target points successfully.  


\subsubsection{Robot Morphologies}
Figure \ref{fig:best5} present the 5 best robots for each method. The morphologies evolved for both tasks are reaching maximum size, having close to 10 modules on average, and are mostly made of hinges with no bricks (except one).
The best robots for point navigation have 3 or 4 limbs made of hinges while those for rotation task only have 2 or 3.

\section{Conclusions and Future Work}
We compared asexual and sexual brain reproduction methods in a morphologically evolving robot system. Since our system also lets `infant' robots optimize their inherited brains by learning, we also had two options regarding the combination of learning and evolution: Darwinian and Lamarckian. Given the two tasks we considered --point navigation and panoramic rotation-- all together we conducted eight experiments. The results show that to achieve the highest task performance, the use of sexual brain reproduction is advisable in the Darwinian system for both tasks. (Although for one task, the two reproduction systems performed similarly.) However, the effect is the opposite in the Lamarckian framework, since asexual reproduction leads to robots with higher fitness. This answers our first research question. Our experiments also show that to maximize task performance, the use of asexual reproduction and the Lamarckian framework is the best choice. This is a novel result that can impact the design of evolutionary robot systems of the future.

With regards to the evolved morphologies, both reproduction methods drive the evolution process towards maximum-size robots composed of many active hinges. The morphologies of the best-performing robots for the same task are similar: they are made of only hinges attached to the core module without bricks. For the rotation task, bodies mainly converged to an "L" shape and for the point navigation task to an "X" shape. We conclude that under the given experimental conditions the morphology of the robot is mainly determined by the tasks, not the brain reproduction methods.

Regarding our third research question, we highlight a difference in the behaviours of the robots that evolved using the two reproduction methods for both tasks. In point navigation, the trajectories of the best robots produced by both reproduction of the Lamarckian framework reached the target points much earlier than the ones from the Darwinian framework where the majority of the best robots evolved using sexual reproduction reach the second target point while only a few of those asexually evolved do so. 

Future work will be directed to test the superiority of asexual brain reproduction and a Lamarckian combination of evolution and learning on more tasks and environments.  

\bibliographystyle{apalike}
\bibliography{references} 
\end{document}